\def\year{2021}\relax
\documentclass[letterpaper]{article} 
\usepackage{aaai21}  
\usepackage{times}  
\usepackage{helvet} 
\usepackage{courier}  
\usepackage[hyphens]{url}  
\usepackage{graphicx} 
\usepackage{natbib}  
\usepackage{caption} 
\usepackage{pbox}
\usepackage{makecell}
\usepackage{multirow}
\usepackage[export]{adjustbox}
\usepackage{mwe}
\usepackage{enumitem}
\usepackage{tablefootnote}
\usepackage{xcolor}
\usepackage[switch]{lineno}  %
\definecolor{mypink1}{rgb}{0.858, 0.188, 0.478}
\definecolor{mypink2}{RGB}{219, 48, 122}

\title{Distilling Localization for Self-Supervised Representation Learning}
\author{
	Nanxuan Zhao$^{1\star}$~~~~~~~~Zhirong Wu$^{2\star}$\blfootnote{Equal Contributions.}~~~~~~~~Rynson W.H. Lau$^{1}$~~~~~~~~Stephen Lin$^{2}$\\
}
\affiliations{
\textsuperscript{\rm 1} City University of Hong Kong ~~~~~~~~~
\textsuperscript{\rm 2} Microsoft Research Asia \\
nanxuanzhao@gmail.com,
rynson.lau@cityu.edu.hk, \{wuzhiron,stevelin\}@microsoft.com \\
\vspace{4pt}
\textcolor{mypink1}{\url{http://nxzhao.com/projects/DiLo_localization/}}
}

\newcommand\blfootnote[1]{%
  \begingroup
  \renewcommand\thefootnote{}\footnote{#1}%
  \addtocounter{footnote}{-1}%
  \endgroup
}

\begin{document}
 

\maketitle

\begin{abstract}

Recent progress in contrastive learning has revolutionized unsupervised representation learning.
Concretely, multiple views (augmentations) from the same image are encouraged to
map to the similar embeddings, while views from different images are pulled apart.
In this paper, through visualizing and diagnosing classification errors, we observe that current contrastive models are ineffective at localizing the foreground object, limiting their ability to extract discriminative high-level features.
This is due to the fact that view generation process considers pixels in an image uniformly.
To address this problem, we propose a data-driven approach for learning invariance to backgrounds.
It first estimates foreground saliency in images and then creates augmentations by copy-and-pasting the foreground onto a variety of backgrounds.
The learning still follows the instance discrimination pretext task,
so that the representation is trained to disregard background content and focus on the foreground.
We study a variety of saliency estimation methods, and find that most methods lead to improvements for contrastive learning.
With this approach (DiLo), significant performance is achieved for self-supervised learning on ImageNet classification, and also for object detection on PASCAL VOC and MSCOCO.
 
\end{abstract}

\section{Introduction}


Visual recognition has been revolutionized by deep learning in the fashion of assembling considerable amounts of labeled data~\cite{deng2009imagenet} and training very deep neural networks~\cite{krizhevsky2012imagenet}. However, collection of supervisory signals, especially at a very large scale, is constrained by budget and time. Due to this, there has been a growing interest in self-supervised and unsupervised learning which do not face this practical limitation.
For high-level visual recognition, previous approaches in self-supervised learning define proxy tasks which do not require human labeling but encode useful priors~\cite{zhang2016colorful,doersch2015unsupervised} for object recognition.
Recent advances in self-supervised contrastive learning rely on
the proxy task of instance discrimination~\cite{dosovitskiy2015discriminative,wu2018unsupervised}, where invariances are encoded and learned from low-level image augmentations such as 
spatial cropping and color jittering.



\begin{figure}[t]
	\centering
	\includegraphics[width=\linewidth]{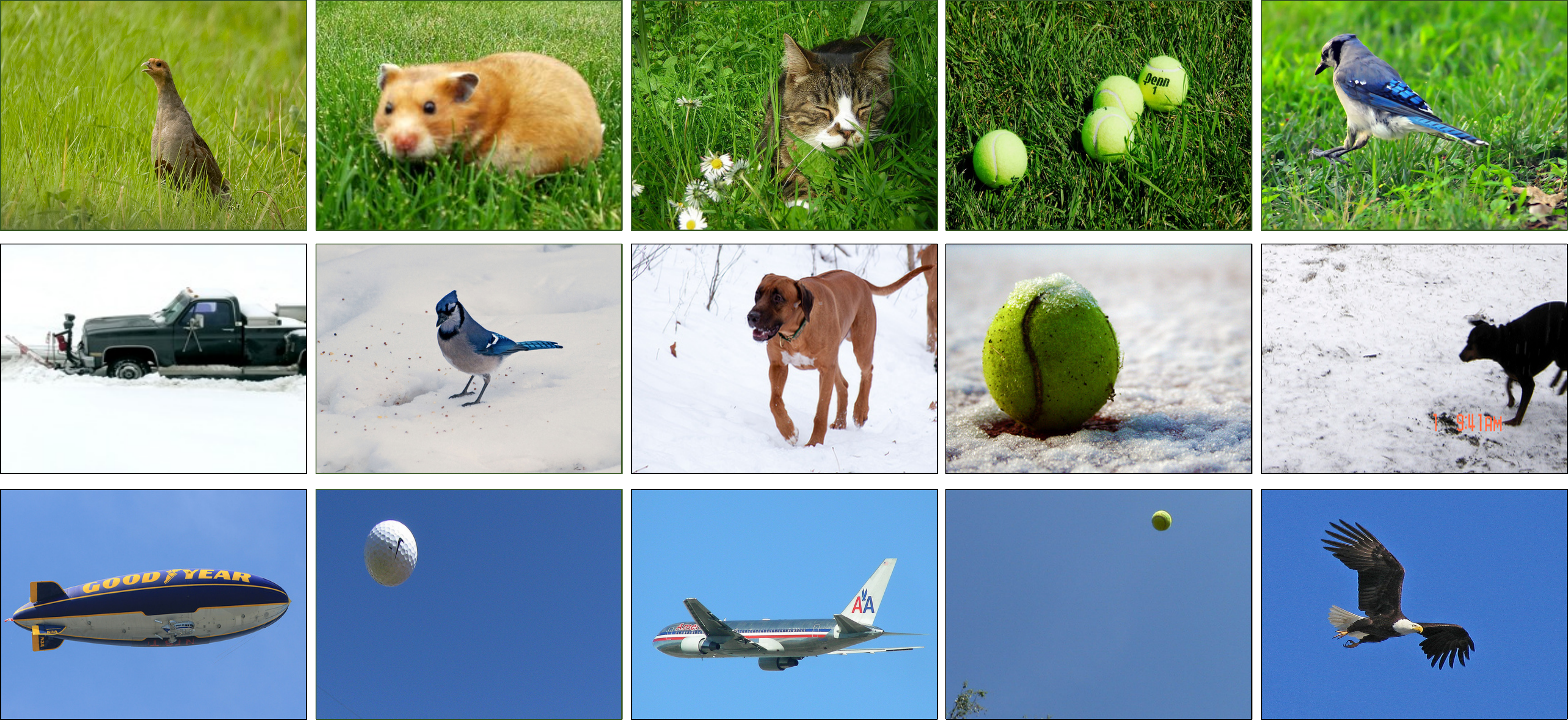}
	\caption{Motivation: for natural images with objects, backgrounds are usually shared across categories, while the distinctive region for determining the object is localized.}
	\label{fig:teaser}
\end{figure}

In this paper, by visualizing and diagnosing errors made by recent self-supervised contrastive models, we identify a strong pattern which is overlooked by prior works. Specifically, we find that current self-supervised models lack the ability to localize foreground objects, and the learned representation can be predominantly determined by background pixels. This is actually unsurprising, as self-supervised learning generally treats each spatial location as equally important, and it is well known that neural networks are prone to ``cheat"~\cite{zhang2016colorful} by taking advantage of unintended information. As a result, a network cannot be expected to discover objects unless it is driven to do so~\cite{arandjelovic2019object}.


In supervised visual recognition, localization has been demonstrated to be a strong by-product of training on image-level labels. Strong object localization performance has been shown using the gradient of the class score in the pixel space~\cite{simonyan2013deep}. It has also been found that adding precise localization information does not bring significant gains for PASCAL object classification when transferred from ImageNet~\cite{oquab2015object}. Moreover, object segments have been estimated using only image-level labels via a class activation mapping method~\cite{zhou2016learning}. As suggested in Figure~\ref{fig:teaser}, we hypothesize that the learning signal that drives localization comes from the category-wise supervisory labels, because background contents (e.g., grass, sky, water) are usually shared among different categories while foreground objects are only salient within the same category.


The gap in the localization ability between self-supervised and supervised models motivates us to explore approaches for distilling localization of self-supervised representations. We study this problem by first estimating a foreground saliency mask for each training image. The training image and its corresponding saliency map are then used to create augmentations by pasting the foreground object onto various backgrounds. During representation learning, we follow recent contrastive representation learning methods using the augmentations for the same object on different backgrounds. This encourages the representation to become invariant to backgrounds, enabling localization of the foreground object.


For generating our augmentations, several saliency estimation methods are examined, including traditional unsupervised techniques~\cite{zhu2014saliency, yan2013hierarchical,wei2012geodesic}, 
and a saliency network~\cite{qin2019basnet}. Our model (DiLo) shows consistent improvements of $2\%-6\%$ over the baselines.
This clearly demonstrates that object recognition benefits from better localization, and that our approach is effective for solving the localization problem.
Due to its better localization ability, 
we also achieve state-of-the-art transfer learning results for object detection on PASCAL VOC and MSCOCO.



In summary, this paper makes the following contributions:


1) A visualization-based study of recent self-supervised contrastive learning models that shows a limited capacity to localize objects.


2) A data-driven method that improves the localization ability of contrastive representation learning, demonstrating its effectiveness on both image classification and object detection transfer tasks.


3) An investigation of different kinds of saliency estimation methods for improving localization, including traditional saliency and network-predicted saliency.

\section{Related Work}


\noindent \textbf{Unsupervised and Self-Supervised Learning.} Unsupervised learning aims to extract semantically meaningful representations without human labels~\cite{de1994learning}. Self-supervised learning is a sub-branch of unsupervised learning which automatically generates learning signals from the data itself. These learning signals have been derived from proxy tasks that involve semantic image understanding but do not require semantic labels for training. These tasks have been based on prediction of color~\cite{zhang2016colorful}, context~\cite{doersch2015unsupervised,pathak2016context}, rotation~\cite{gidaris2018unsupervised}, and motion~\cite{pathak2017learning}.  Auto-encoders~\cite{vincent2008extracting} and GANs~\cite{goodfellow2014generative,donahue2019large} have also shown promising results for representation learning through reconstructing images.

Contrastive learning is another promising direction of work for self-supervised learning.
It achieves invariances in a data-driven fashion by image augmentations. Exemplar CNN~\cite{dosovitskiy2015discriminative} and instance discrimination~\cite{wu2018unsupervised} create augmentations of an image through changes in color, spatial location and scale.
PIRL~\cite{misra2020self} and CPC~\cite{oord2018representation} formulate contrastive learning in image patches.
CMC~\cite{tian2019contrastive} considers explicit modeling of different views.
MoCo~\cite{he2019momentum} and SimCLR~\cite{chen2020simple} scale contrastive learning by momentum encoders and large batch sizes. 
Our paper is in line with these works, and we propose a non-trivial 
augmentation for distilling localization information.

\vspace{3pt}
\noindent \textbf{Saliency Estimation.}
Saliency estimation refers to the task of estimating the locations of interesting objects consistent with human perception. For learning saliency, datasets~\cite{mit-saliency-benchmark} have been collected by tracking eye fixations over an image.
Later works usually consider saliency as the full foreground object. 

Previous non-learning based approaches~\cite{zhu2014saliency,yang2013saliency} rely on handcrafted features and use priors to find salient object regions.
Useful priors include background priors~\cite{han2014background}, color contrast priors~\cite{cheng2014global}, and objectness~\cite{jiang2013salient}.
Deep supervised methods~\cite{qin2019basnet} train a segmentation network to regress the foreground mask, outperforming all non-learning based methods.
Recent research on saliency estimation also explores unsupervised learning
methods.
It integrates multiple non-learning based methods into a noise optimization framework~\cite{zhang2017supervision}, showing results that are on par with supervised methods.

In a network, the salient region corresponds to pixels that fire for the classification decision. Previous works study this in both the input space via gradient visualization~\cite{simonyan2013deep} and the output space via activation mapping~\cite{zhou2016learning}. A prior work~\cite{zhou2014object} also finds the salient region by optimizing a minimal region that determines the classification response. 

\begin{figure*}
	\centering
	\includegraphics[width=\linewidth]{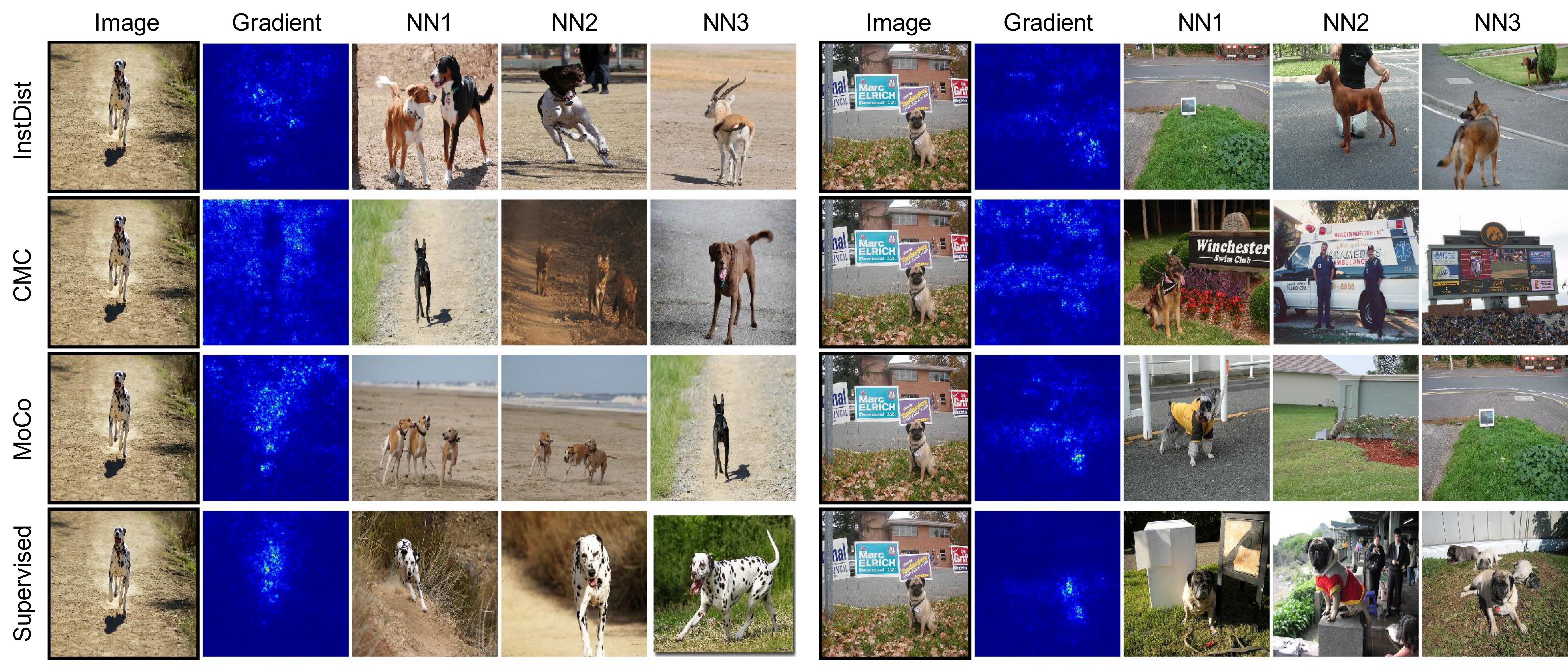}
	\caption{\small Visualizing and analyzing the error patterns of self-supervised contrastive models. Given an input for each model, we visualize its top-3 nearest neighbors in the embedding space, as well as the gradient on the pixel space with respect to the classification signal. Compared with the supervised model, which is able to localize the salient objects, self-supervised models (InstDisc, CMC, MoCo) look holistically over the image and are prone to distraction by backgrounds.
	More visualizations are available in the appended supplement.}
	\label{fig:diag}
\end{figure*}

\noindent \textbf{Copy-and-paste for Visual Recognition.}
Several works create data in a copy-and-paste fashion for visual recognition.
A key insight of such an approach is that data being generated may not look realistic, but the trained model generalizes surprisingly well to real data. For example,
Flying Chairs~\cite{dosovitskiy2015flownet} renders chairs onto various backgrounds to generate data for optical flow estimation.
Cut-paste-learn~\cite{dwibedi2017cut} randomly puts household object instances in an indoor environment for instance detection and segmentation.
Instaboost~\cite{fang2019instaboost} spatially shifts the foreground objects as a means of data augmentation for instance segmentation.
Copy-pasting GAN~\cite{arandjelovic2019object} uses the copy-and-paste idea to discover objects in an unsupervised manner. However, their experiments are performed on toy examples, such as discovering artificial boxes. Moreover, they do not show how discovering objects may help recognition. 
Our work follows this path, but in contrast to these previous works our method is targeted to self-supervised representation learning. We note that our augmented images are extremely unrealistic, but provide useful information for learning a recognition model.

\noindent \textbf{Image Augmentations. } Data augmentation plays a key role in visual recognition. Recent works devise handcrafted augmentations~\cite{devries2017improved} or learning-based methods~\cite{cubuk2019autoaugment,ratner2017learning} to boost representation learning especially in semi-supervised learning. Our copy-paste augmentation is the first introduced for self-supervised learning. From it, we seek to gain further understanding about the ineffective localization problem in self-supervised learning.

\section{Revisiting Contrastive Learning}


Our work builds on recent contrastive learning methods for unsupervised learning, where most work follow the pretext task of instance discrimination. The algorithm first generates image augmentations in the spatial domain, scale space, and color space and then it encourages augmentations of the same image to have similar feature embeddings, and augmentations of different images to have dissimilar embeddings.


Let $x$ denote the image and $\mathbf{v} = f(x)$ be the feature embedding, where $f(\cdot)$ is the embedding function implemented as a convolutional neural network. Let $\tilde{x} = \mathcal{T}(x)$ represent an augmentation for image $x$, where $\mathcal{T}$ is a random augmentation function. The probability of the augmentation $\tilde{x}$ to be classified as the $i$-th identity is expressed as
\begin{equation}\label{eqn:np_softmax}
	P(i | \tilde{\mathbf{x}}) =
	\frac{\exp \left( \mathbf{v}_i^T \tilde{\mathbf{v}} / \tau \right)}
	{\sum_{j=1}^n \exp \left( \mathbf{v}_j^T \tilde{\mathbf{v}} / \tau \right)},
\end{equation}
where $\tau$ is a temperature parameter and $n$ is the total number of images in the dataset.
$\tilde{\mathbf{v}}=f(\tilde{\mathbf{x}}), \mathbf{v}_i=f(\mathbf{x}_i)$ are the embeddings for image $x_i$ and $\tilde{x}$. The learning objective is to minimize the negative log-likelihood over the dataset:
\begin{equation} \label{eqn:obj0}
J(\mathbf{\theta})
=  - \sum_{i=1}^n \log P(i | f_{\mathbf{\theta}}(\mathcal{T}(x_i))).
\end{equation}


Recent self-supervised learning methods such as InstDisc~\cite{wu2018unsupervised}, CMC~\cite{tian2019contrastive}, MoCo~\cite{he2019momentum}, SimCLR~\cite{chen2020simple} all share a similar formulation. The effectiveness of such an approach for unsupervised learning strongly relies on the types of augmentations $\mathcal{T}(\cdot)$, i.e., image transformation priors that do not change object identity. In Table~\ref{table:invariance}, we summarize the role of data-driven augmentations for both a typical self-supervised MoCo ResNet50 model~\cite{he2019momentum} and the supervised model. We gradually add each type of transformation to the set of augmentations. The performance is measured on the ImageNet validation set of 1000 classes, and evaluated by linear classifiers. 


We find that the unsupervised representation gains much more classification accuracy from the augmentations than the supervised representation. This indicates that the priors present in the augmentations strongly overlap with the modeling cues from semantic labels. Adding intense color jittering improves the unsupervised representation but hurts the supervised representation. This suggests that the color jitter prior expands beyond the original data distributions.
Nevertheless, adding a prior that only partially relates to semantics improves self-supervised learning significantly. 

\begin{table}[t]
	\setlength{\tabcolsep}{8pt}
	\centering
	\renewcommand\arraystretch{1.1}
	\caption{A comparison study of the role of data augmentations for learning self-supervised and supervised representations. Please refer to the main text for details.}
	\vspace{0.1cm}
	\begin{tabular}{l|c|c}
		\Xhline{3\arrayrulewidth}
		Augmentations & Self-Supervised & Supervised \\
		\Xhline{3\arrayrulewidth}
	    + Flipping & 6.4 & 70.9 \\
	    \hline
	    + Spatial Scale Crop & 40.4 & 77.5 \\
	     \hline
	     + Color Jitter & 56.9 &  77.4 \\
	     \hline
	     + Random Gray & 60.6  & 77.7 \\
		\Xhline{3\arrayrulewidth}
	\end{tabular}
	\label{table:invariance}
\end{table}

\section{Visualizing / Diagnosing Contrastive Learning}

\begin{figure*}[t]
	\centering
	\includegraphics[width=0.8\linewidth]{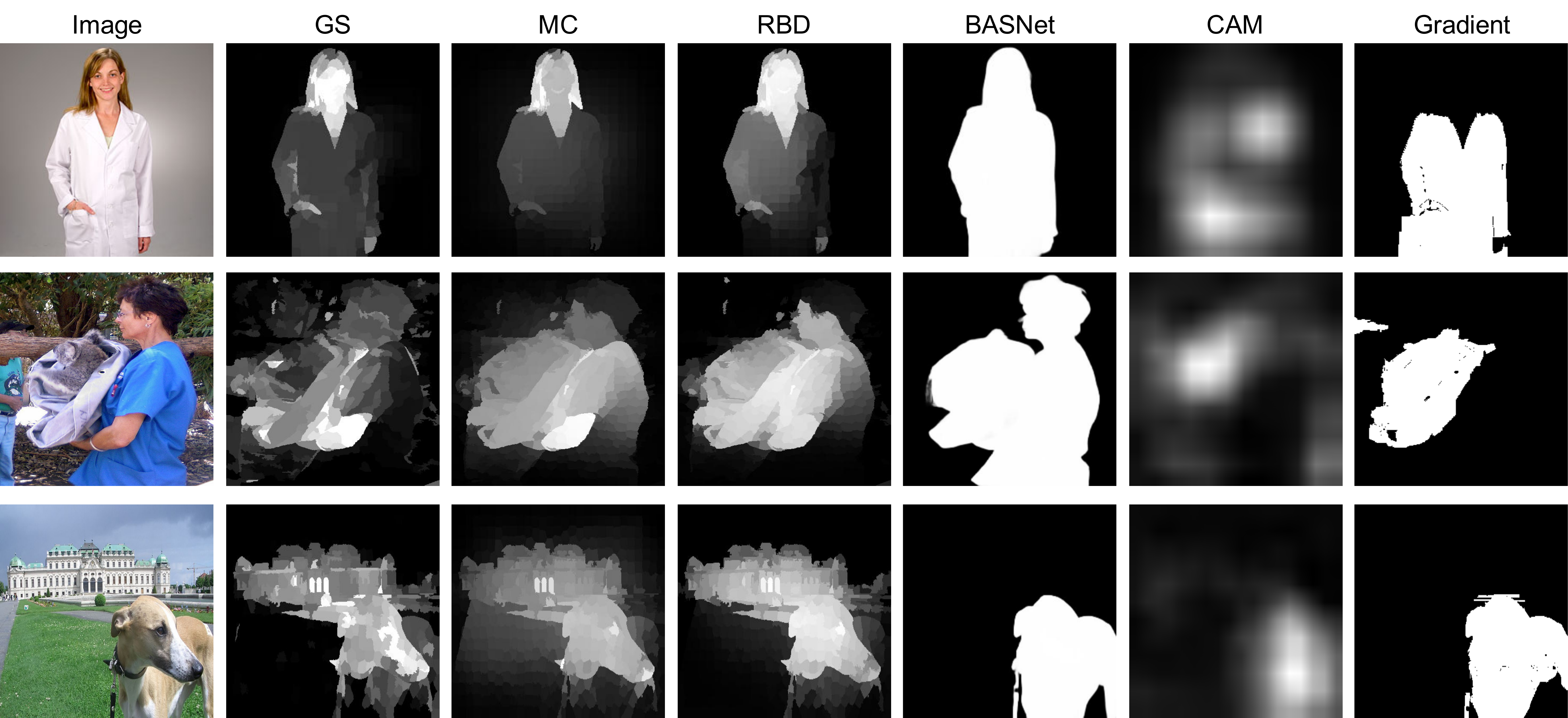}
	\caption{\small Examples of saliency estimations methods. We show 6 saliency estimations, including traditional methods (GS~\cite{wei2012geodesic}, MC~\cite{jiang2013saliency}, RBD~\cite{zhu2014saliency}), a network predicted saliency BASNet~\cite{qin2019basnet}, and class-specific methods visualized from a pretrained network (CAM~\cite{zhou2016learning}, Gradient~\cite{simonyan2013deep}). }
	\label{fig:saliency}
\end{figure*}




A variety of methods have been presented for visualizing the behavior of supervised convolutional neural networks, based on deconvolution~\cite{zeiler2014visualizing}, class-specific gradients~\cite{simonyan2013deep}, and class activation mapping~\cite{zhou2016learning, selvaraju2017grad}. However, there is little work on visualizing and analyzing the error patterns of self-supervised models, particularly for understanding the relationship between the proxy task and the semantic labels.

In the following, we visualize some representative contrastive learning models with a focus on understanding the salient regions when self-supervised networks make wrong predictions.


\vspace{3pt}

\noindent \textbf{Visualization Methods.} We adapt two visualization methods to our objective.
\begin{itemize} 

\vspace{-0.1cm}
 \item Nearest Neighbors. A straightforward way to diagnose what a feature has learned is to find the nearest neighbors in the feature space. By identifying patterns on what draws neighbors close to each other, we gain insights about what the features represent.

\vspace{-0.1cm}
 \item Class-specific gradients. The magnitude of class-score gradients in the pixel space provides information about how important the pixels are for classification. This approach has proven to be strong for weakly-supervised object localization~\cite{simonyan2013deep}. Since self-supervised models do not have classifiers for objects, we train a linear classifier on top of the extracted features. Then we do back-propagation through the linear classifier and the rest of the self-supervised network to calculate the gradients in the pixel space.

\end{itemize}

\noindent \textbf{Investigated Models.}
We examine three self-supervised models, including InstDist, CMC, and MoCo.

\begin{itemize}
\vspace{-0.1cm}
 \item InstDist~\cite{wu2018unsupervised} treats each individual instance as a class and learns a representation by non-parametric  classification with a memory bank implementation.
 \vspace{-0.1cm}
\item CMC~\cite{tian2019contrastive} explicitly decouples an image to two views, namely the lightness and color channels. Learning follows to maximize the mutual information between views.
\vspace{-0.1cm}
\item MoCo~\cite{he2019momentum} follows InstDist and further proposes a momentum encoder to fix the consistency between positives and a queue-based memory for scalability.
\end{itemize}

\noindent \textbf{Error Patterns.}
Figure~\ref{fig:diag} illustrates our major findings. We observe that for a considerable number of error cases, the similarity between a query and its nearest neighbors exists mainly in their backgrounds. Gradient-based saliency visualization confirms such findings, as the salient regions for self-supervised models are spread across the background instead of the foreground. For comparison, we also show the corresponding results for the supervised models, which instead show similarities among the foregrounds.


Since these self-supervised methods rely heavily on augmentations to learn invariances, and these augmentations treat foreground and background pixels equally, thus they do not enforce a loss that drives the model to discover objects. This lack of localization ability calls for salient region modeling in self-supervised learning.

\section{DiLo: Distilling Localization via \\ Background Invariance}

Our goal is to learn a representation from which the foreground object can be automatically localized, such that discriminative regions can be focused on to improve recognition. We propose to distill the ability for object localization by learning invariance against the background. We first describe methods to extract foreground regions by saliency estimation, and then introduce our background augmentations by copy-and-paste operations.

\subsection{Saliency Estimation}
\label{sec:saliency}


In distilling localization ability for self-supervised methods, our approach first estimates saliency masks. A saliency mask should depict the regions most relevant for classifying the object. Typically, it coincides with the foreground object region, as indicated by most saliency datasets~\cite{wang2017}.


Note that recent research on unsupervised saliency estimation has shown promising progress. However, these models~\cite{zhang2018deep,nguyen2019deepusps} heavily rely on ImageNet and semantic segmentation pretraining, which violates our unsupervised experimental protocols. We avoided these methods in this paper, and instead consider the following techniques. 


\vspace{3pt}

\noindent \textbf{Traditional Methods.} 
Traditional saliency estimation methods use handcrafted features, and rely on
priors and heuristics to find the dominant object in an image. Useful priors include the background prior (pixels on the image border are more likely to be background) and the color contrast prior (edges with high contrast tend to belong to the foreground). We investigate several high-performing methods: RBD~\cite{zhu2014saliency}, MC~\cite{jiang2013saliency}, and GS~\cite{wei2012geodesic}.

\vspace{3pt}


\noindent \textbf{Saliency Networks.} Recent methods for saliency estimation commonly employ deep learning on annotated saliency datasets~\cite{wang2017}. These deep models outperform traditional methods by a large margin. A state-of-the-art saliency network BASNet~\cite{qin2019basnet} is included in the investigation, and it is trained on a modest amount of 10K images from scratch.

\vspace{3pt}





\noindent \textbf{Class-specific Saliency.} The aforementioned methods estimate saliency as foreground object regions. However, it is not clear that this represents the discriminative part of an image (e.g., only the face of a person may be important for recognizing humans). 
To keep the problem open, we also compare with CAM~\cite{zhou2016learning} and a gradient-based method~\cite{simonyan2013deep} through class-specific visualizations. For~\cite{simonyan2013deep}, we convert the gradients to a mask using a segmentation algorithm~\cite{gulshan2010geodesic}.

\vspace{2pt}

\noindent \textbf{Summary.}  Figure~\ref{fig:saliency} shows examples of the saliency visualizations. Traditional methods are seen to be noisy, while network-produced saliency is much cleaner. It can be noticed that class-specific saliency from a pretrained network tends to be more compact around discriminative regions. This indicates that the use of full foreground saliency may not be ideal.


\begin{figure}[t]
	\centering
	\includegraphics[width=\linewidth]{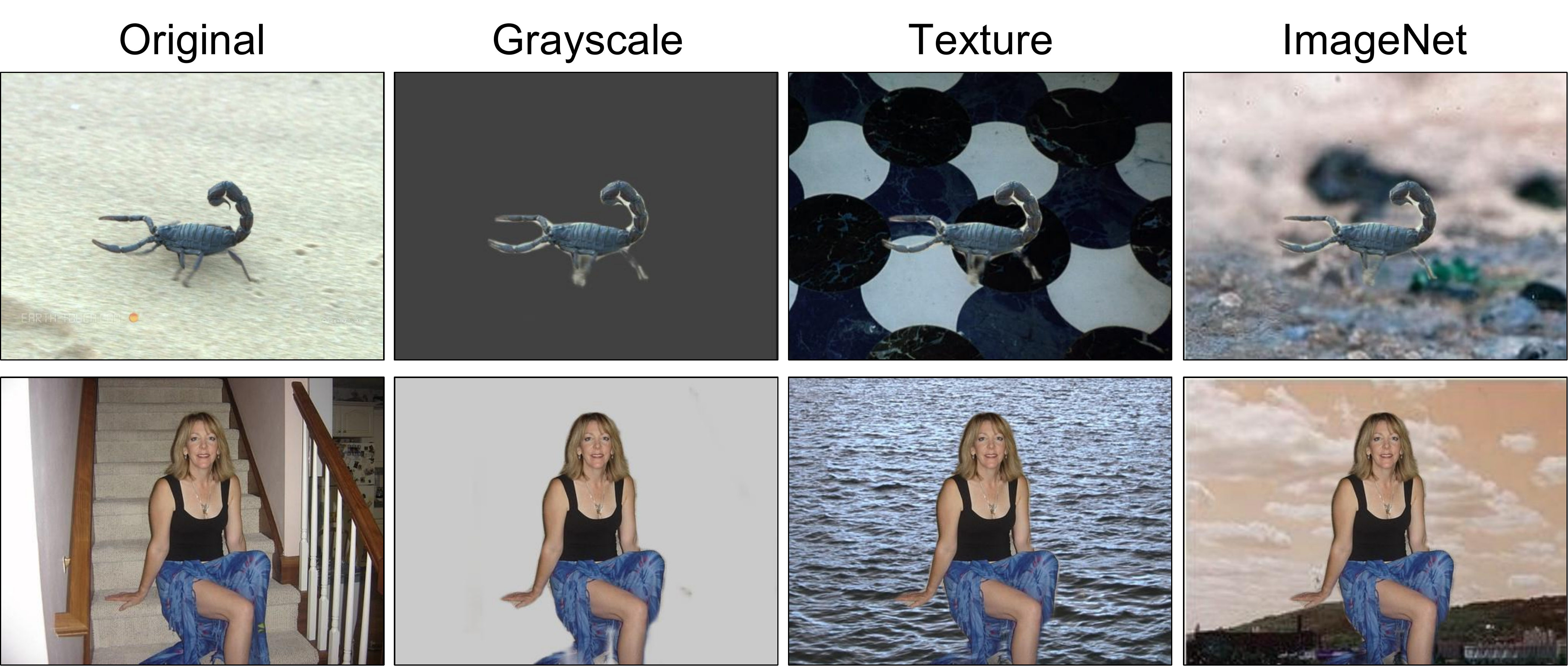}
	\caption{\small Generated copy-paste augmentations using three kinds of background images.}
	\label{fig:copy_paste}
\end{figure}

\subsection{Copy-and-paste for Background Augmentation}

Based on the previous findings, we propose to copy the foreground object estimated from the saliency methods in prior section~\ref{sec:saliency}, and paste that onto various backgrounds as a means of data-driven augmentation for learning localization.

\noindent \textbf{Background Datasets.}
For this augmentation, we ablate three types of backgrounds.
\begin{itemize}
\vspace{-3pt}
    \item Homogeneous grayscale images with a random grayscale level.
    \item Texture images from the MIT Vision Texture dataset~\cite{medialab1995vistex}.
    \item Image crops from ImageNet which have no saliency response using RBD~\cite{zhu2014saliency}.
   
\end{itemize}
 Figure~\ref{fig:copy_paste} shows copy-and-pasted examples using various background images.
 
 \vspace{3pt}
\noindent \textbf{Blending.}
For pasting, we examine three techniques: directly copying the foreground object onto the background, copying with Gaussian blending on the object borders, and a mixture of the two approaches.

 \vspace{3pt}
\noindent \textbf{Accounting for Context.}
Context plays an important role in recognizing objects~\cite{torralba2003contextual}. Though the surrounding context of an object may not be the most discriminative region for recognition, it may help to prune the set of candidates. For example, a tree is unlikely to be completely  encompassed by sky. To account for this during augmentation, we set a probability of keeping the original full image without copy-and-paste augmentation. 

\vspace{3pt}
\noindent \textbf{Integrating other Augmentations.} Since copy-paste augmentation is orthogonal to other previous augmentations, i.e. random scaling, cropping, color jittering, the order of copy-paste augmentation with respect to other augmentations does not matter.
In our implementation, we first run copy-paste augmentation to replace the background, and then perform other augmentations.

\section{Experiments}

We conduct a series of experiments on model designs for self-supervised representation learning
and their transfer learning abilities.



\begin{table*}[t]
 %
 \label{exp:abl}
 \caption{ \small Ablation studies for investigating copy-and-pasting augmentations: (a) on various saliency estimation methods (b) on controlling the ratio of using copy-and-pasting augmentation (c) on various background images (d) on blending options.}
 \vspace{-6pt}
 \begin{minipage}{0.30\textwidth}
  \centering
  \small
  \setlength{\tabcolsep}{4pt}
  (a)
\begin{tabular}{c|cc|c|c}
\Xhline{2\arrayrulewidth}
Saliency      & $F_{\beta}$ & MAE  &  Acc & $\Delta$  \\ \Xhline{2\arrayrulewidth}
MoCo& -           & -     & 60.6 &  -       \\ \Xhline{2\arrayrulewidth}
GS            & 0.557       & 0.173 &  62.7 & +2.1  \\
MC            & 0.627       & 0.186 &  62.1 &  +1.5  \\
RBD           & 0.630       & 0.144 &  62.8 &  +2.2    \\
BASNet        & 0.805       & 0.056 &  65.0 &  +4.4 \\ \Xhline{2\arrayrulewidth}
\end{tabular}
\\[0pt]
  \label{exp:abl_sal}
 \end{minipage}%
  \begin{minipage}{0.23\textwidth}
  \centering
   \small
  \setlength{\tabcolsep}{5pt}
  (b)
  \begin{tabular}{c|c|c}
\Xhline{2\arrayrulewidth}
Aug Ratio & Linear & $\Delta$  \\
\Xhline{2\arrayrulewidth}
MoCo & 60.6 & - \\
\Xhline{2\arrayrulewidth}
30\% &  62.8 & +2.2\\
50\% & 62.2 &  +1.6\\
70\% & 61.6 &  +1.0 \\
100\% & 47.6 &  -13.0 \\
\Xhline{2\arrayrulewidth}
\end{tabular}
\\[0pt]
  \label{exp:abl_augment_amount}
 \end{minipage}%
  \begin{minipage}{0.23\textwidth}
  \centering
   \footnotesize
  \setlength{\tabcolsep}{5pt}
 (c)
  \begin{tabular}{c|c|c}
\Xhline{2\arrayrulewidth}
Background & Linear & $\Delta$  \\
\Xhline{2\arrayrulewidth}
MoCo & 60.6 & - \\
\Xhline{2\arrayrulewidth}
Texture & 60.6 & +0.0\\
Imagenet & 62.1 & +1.5\\
Grayscale & 62.8 &  +2.2\\
\Xhline{2\arrayrulewidth}
\end{tabular}
  \label{exp:abl_background}
 \end{minipage}%
  \begin{minipage}{0.23\textwidth}
  \centering
   \footnotesize
  \setlength{\tabcolsep}{5pt}
 (d)
  \begin{tabular}{c|c|c}
\Xhline{2\arrayrulewidth}
Blending & Linear & $\Delta$  \\
\Xhline{2\arrayrulewidth}
MoCo & 60.6  & - \\
\Xhline{2\arrayrulewidth}
No blend & 62.4 & +1.8\\
Gaussian & 62.5 & +1.9\\
Mix &  62.8 & +2.2\\
\Xhline{2\arrayrulewidth}
\end{tabular}
  \label{exp:abl_blend}
 \end{minipage}%
  \vspace{-8pt}
\end{table*}

\subsection{Ablation Study}

In this section, we first validate our data-driven approach of distilling localization through
a  series of ablation experiments for image classification on ImageNet.

\noindent \textbf{Baseline Settings.}
Due to its state-of-the-art performance, we largely follow MoCo~\cite{he2019momentum} settings as our baseline for ablation.
Specifically, we use a temperature $\tau = 0.07$ in Eqn.~\ref{eqn:np_softmax}, and an embedding dimension of $D=128$ for each image. A memory queue~\cite{he2019momentum} of size $k=65536$ negatives is used to accelerate discrimination.
Training takes 200 epochs with an initial learning rate of $0.03$ that is decayed $1/10$ at epochs 120 and 160.
All models are trained using the ResNet50 architecture and reported on the ImageNet validation set.
Performance is evaluated by the linear readoff on the penultimate layer features. The optimization takes 100 epochs and starts with a learning rate of 30 that is decayed every 30 epochs.


\noindent \textbf{A naive approach.}
First of all, to demonstrate the necessity of a data-driven approach, we consider a naive approach that pools the final layer features by masking according to saliency.
With this, the performance decreases sharply by $19\%$,
possibly because the model loses too much context.
Moreover, by masking out the features, the model is still unable to localize the discriminative regions automatically.

\vspace{2pt}
\noindent \textbf{Saliency Estimation.}
In Table~\ref{exp:abl_sal} (a), we examine several class-agnostic saliency estimation methods. All of them are found to improve performance, even the noisy traditional approaches RBD~\cite{zhu2014saliency}, MC~\cite{jiang2013saliency} and GS~\cite{wei2012geodesic}. RBD improves the performance by $2.2\%$ and the saliency network BASNet by $4.2\%$. The supervised BASNet~\cite{qin2019basnet} is trained on the DUTS dataset~\cite{wang2017} from scratch with 10,053 training images, which is less than $1\%$ of ImageNet.
This indicates potential room for developing better unsupervised saliency approaches.
In Table~\ref{exp:abl_sal}, we find a correlation between the saliency performance on the saliency benchmark (by $F_{\beta}$ and MAE on DUT-OMRON dataset~\cite{yang2013saliency}) and the self-supervised representation learning. Better saliency translates to better representations.




\vspace{2pt}
\noindent \textbf{Background Images.}
We ablate the use of various background images in Table~\ref{exp:abl_background} (c).
Texture backgrounds improve the performance very marginally.
This is possibly because textured images in the dataset~\cite{mit-saliency-benchmark} are outside of the ImageNet distribution.
Homogeneous grayscale backgrounds and ImageNet backgrounds perform similarly well.

\vspace{2pt}
\noindent \textbf{Amount of Augmentation.}
During dataloading, we only randomly add copy-and-paste augmentations with a probability ratio.
We ablate the ratio in Table~\ref{exp:abl_augment_amount} (b).
With only $30\%$ to $50\%$ of images receiving copy-and-pastes, we significantly improve the performance by $2\%-4\%$.
Always using the copy-and-paste augmentation hurts performance.

\vspace{2pt}
\noindent \textbf{Blending Options.}
When copy-and-pasting an object to a background, blending has proven to be important
for object detection~\cite{dwibedi2017cut}. In our study in Table~\ref{exp:abl_blend} (d), blending appears to improve the performance minorly about $0.4\%$. This difference is possibly because detection requires realistic boundaries, which prevents the network from taking shortcuts,
while for classification, boundary cheating is not as significant.


\begin{figure}[t]
	\centering
	\includegraphics[width=\linewidth]{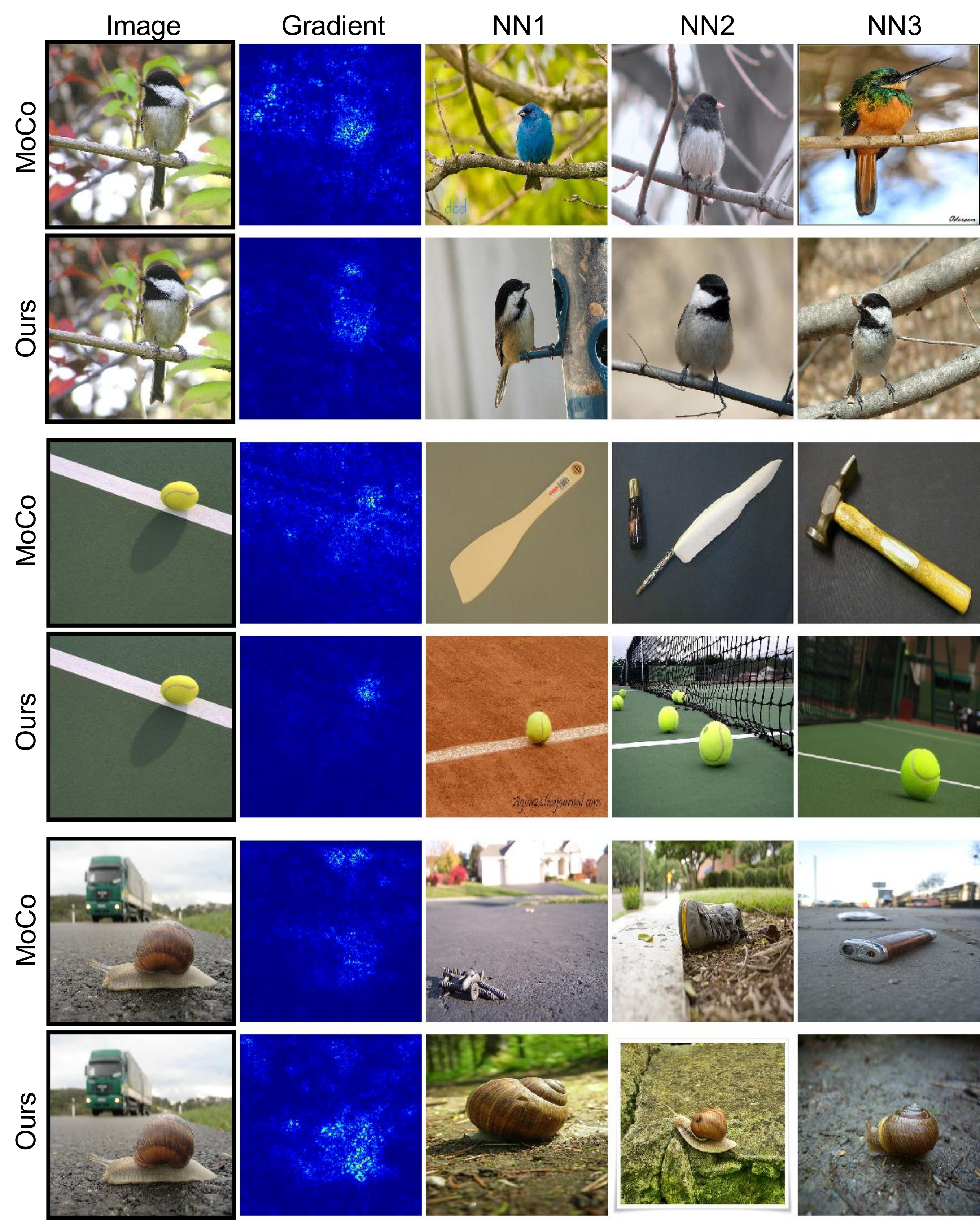}
	\caption{\small Successful examples where our model outperforms our baseline. The improvement is due to better localization and background invariance.}
	\label{fig:visualization_success}
\end{figure}

\begin{figure}[t]
	\centering
	\includegraphics[width=\linewidth]{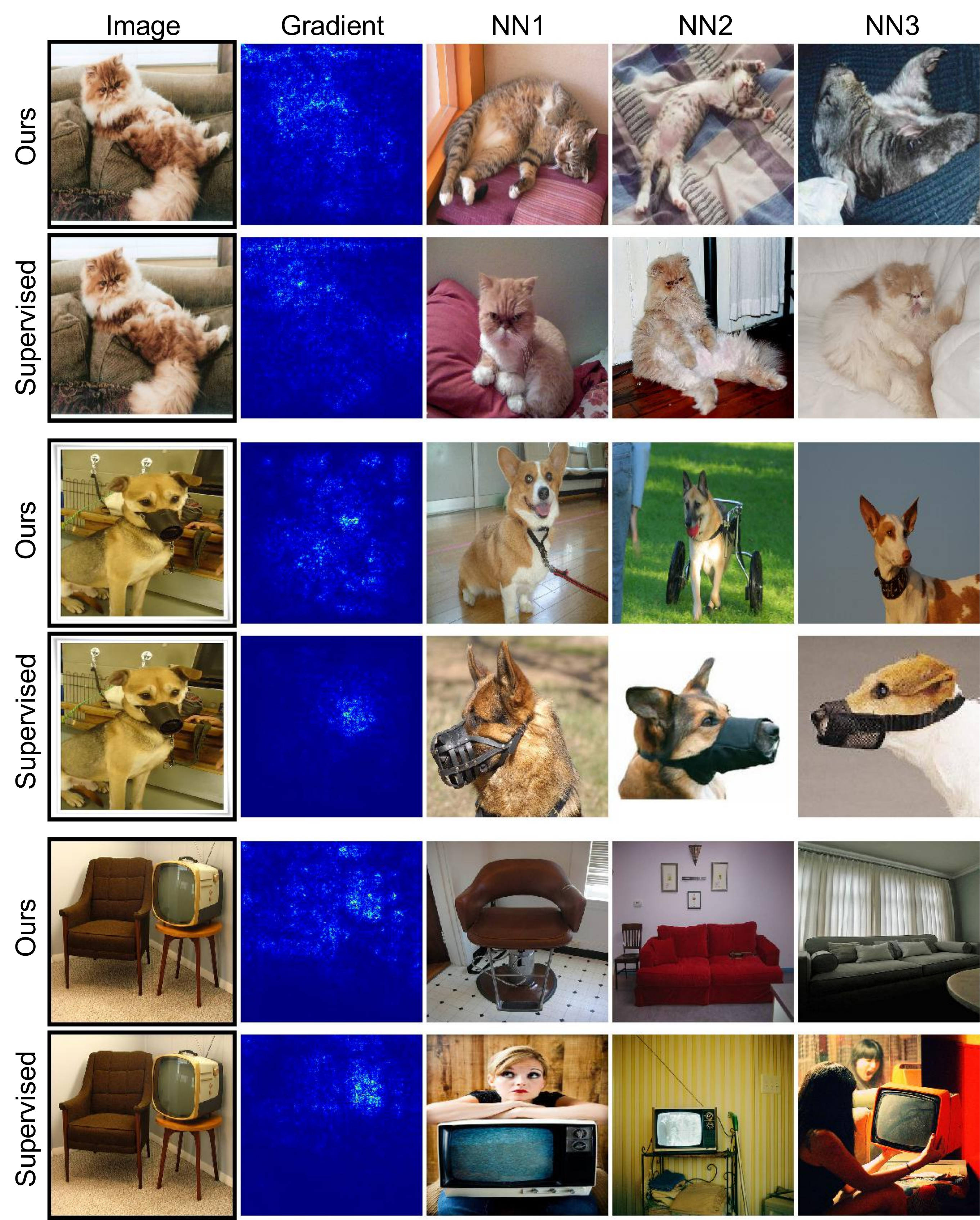}
	\caption{\small Failures where our model underperforms the supervised model. The model finds it difficult when multiple objects appear in the image, or the object is of a fine-grained category.}
	\label{fig:visualization_failure}
\end{figure}

\vspace{2pt}
\noindent \textbf{Visualizations.}
In Figure~\ref{fig:visualization_success} and Figure~\ref{fig:visualization_failure}, we visualize examples where our model outperforms the baseline, as well as some failure cases.
For all the successful cases, our salient region on the gradient and the nearest neighbors focus on the
discriminative object, while the baseline approach is distracted by the background.
This validates the claim that our data-driven augmentation drives the model to learn to automatically localize the object. Such localization leads to better recognition performance.

For the failure cases, we compare our model with the supervised model.
We find that there are two error patterns. First, multiple objects appear in a single image, and our model makes wrong decisions on where to focus. Second, the testing image is of a fine-grained class, too difficult to recognize without labels. 

\begin{table}
  \centering
  \small
  \setlength{\tabcolsep}{8pt}
\caption{Distilling localization on various contrastive representation learning models for ImageNet classification.}
  \begin{tabular}{c|c|c|c}
\Xhline{2\arrayrulewidth}
Methods & Original & DiLo-RBD &  DiLo-BasNet \\
InstDist & 56.5 & 59.3 & 62.9 \\
CMC & 63.4 & 65.0 & 66.9 \\
MoCo-v1 & 60.6 & 62.8 & 65.0 \\
MoCo-v2 & 67.5 & 67.9 & 69.2 \\
\Xhline{2\arrayrulewidth}
\end{tabular}

\label{exp:imagenet_all}

\end{table}

\subsection{Transfer Learning Results}

We evaluate the transfer learning ability of our model on object recognition, and object detection benchmarks, and compare with the state-of-the-art methods.


\vspace{2pt}
\noindent \textbf{ImageNet Classification.}
We conduct a plug-and-play of DiLo into existing contrastive learning frameworks. Methods being investigated include InstDist~\cite{wu2018unsupervised}, CMC~\cite{tian2019contrastive}, MoCo~\cite{he2019momentum} and MoCo-v2~\cite{chen2020improved}.
In Table~\ref{exp:imagenet_all}, DiLo consistently improves image classification on all baselines.
Foreground masks estimated from BasNet are more beneficial than RBD.
The results demonstrate that DiLo is orthogonal to prior contrastive learning works.

\begin{table}[t]
\centering
\footnotesize
\setlength{\tabcolsep}{5pt}
\renewcommand\arraystretch{1.1}
\caption{Transfer learning for object detection on VOC 0712. We present the gap to ImageNet supervised pre-training in the brackets for reference. All numbers are the averages of three independent runs.}
\label{exp:detection}
\begin{tabular}{l|cc|cc|cc}
\Xhline{2\arrayrulewidth}
Method & \multicolumn{2}{c|}{ $AP$} & \multicolumn{2}{c|}{$AP_{50}$} & \multicolumn{2}{c}{$AP_{75}$} \\ \Xhline{2\arrayrulewidth}
Supervised & 53.5 & - & 81.3 & -  & 58.8 & -  \\
MoCo & 55.9 & (+2.4) & 81.5 & (+0.2)  & 62.6 & (+3.8)\\
DiLo-RBD & 56.5 & (+3.0)& 81.9 & (+0.6)  & 63.3 & (+4.5)\\
DiLo-BasNet & 56.9 & (+3.4) & 82.1 & (+0.8)  & 64.1 & (+5.3) \\ \Xhline{2\arrayrulewidth}
\end{tabular}
\end{table}

\vspace{2pt}
\noindent \textbf{Object Detection on PASCAL VOC.}
We transfer our pretrained model to object detection
by finetuning it on PASCAL VOC 2007+2012 trainval and evaluating on the VOC 2007 test set.
Following the state-of-the-art method MoCo~\cite{he2019momentum}, we use the exact same training protocol to finetune the Faster R-CNN with a Res50-C4 backbone as with the supervised counterpart. A critical BN layer is added after the conv5 stage in the box prediction head. During training, we finetune all layers with synchronized BN.
The finetuning takes 9k iterations.
Results are summarized in Table~\ref{exp:detection}.
DiLo-RBD and DiLo-BasNet consistently outperform supervised baseline and MoCo on all metrics, especially on $AP_{75}$ which heavily reflects localization ability.

\vspace{2pt}
\noindent \textbf{Object Detection and Instance Segmentation on COCO.}
We transfer the pretrained DiLo model for object detection and instance segmentation on MSCOCO by finetuning it with the Mask-RCNN Res50-FPN pipeline using the Detectron2 codebase. Finetuning takes the default 1x schedule. DiLo-RBD and DiLo-BasNet consistently outperform the MoCo baseline with good margins on both detection and segmentation.   

\begin{table}[t]
\centering
\footnotesize
\setlength{\tabcolsep}{3pt}
\renewcommand\arraystretch{1.1}
\caption{Transfer learning for object detection and instance segmentation on COCO. Model is finetuned with Mask-RCNN ResNet50-FPN pipeline and 1x schedule.}
\label{exp:detection}
\begin{tabular}{l|ccc|ccc}
\Xhline{2\arrayrulewidth}
Method &  $AP^{bb}$& $AP^{bb}_{50}$ & $AP^{bb}_{75}$ & $AP^{mk}$ & $AP^{mk}_{50}$ & $AP^{mk}_{75}$ \\ \Xhline{2\arrayrulewidth}
Supervised & 39.7 &  59.5 & 43.3 & 35.9 & 56.6 & 38.6 \\
MoCo & 39.4 & 59.1 & 42.9 & 35.6 & 56.2 & 38.0\\
DiLo-RBD & 39.8 & 59.5 & 43.3 & 36.0 & 56.7 & 38.6\\
DiLo-BasNet & 40.1 & 60.0 & 44.0 & 36.3 & 56.8 & 39.0\\ \Xhline{2\arrayrulewidth}
\end{tabular}
\end{table}

\section{Conclusion}
In this work, we identified a strong error pattern among
self-supervised models in their failure to localize foreground objects.
We then propose a simple data-driven approach to distill localization via learning invariance against backgrounds.
We achieve strong results on ImageNet classification and its transfer performance for object detection on PASCAL VOC 2007.
The improvements achieved suggest that the localization problem for self-supervised representation learning is prevalent.
However, our method may not be the ideal way to solve this localization problem.
We are interested in finding a clever ``proxy task" which can help 
distill such localization abilities.

\newpage
\bibliography{egbib}

\appendix
\clearpage
\title{Supplementary Materials}
\renewcommand{\thesection}{A\arabic{section}}

 \begin{figure}[h!]
 	\centering
 	\includegraphics[width=\linewidth]{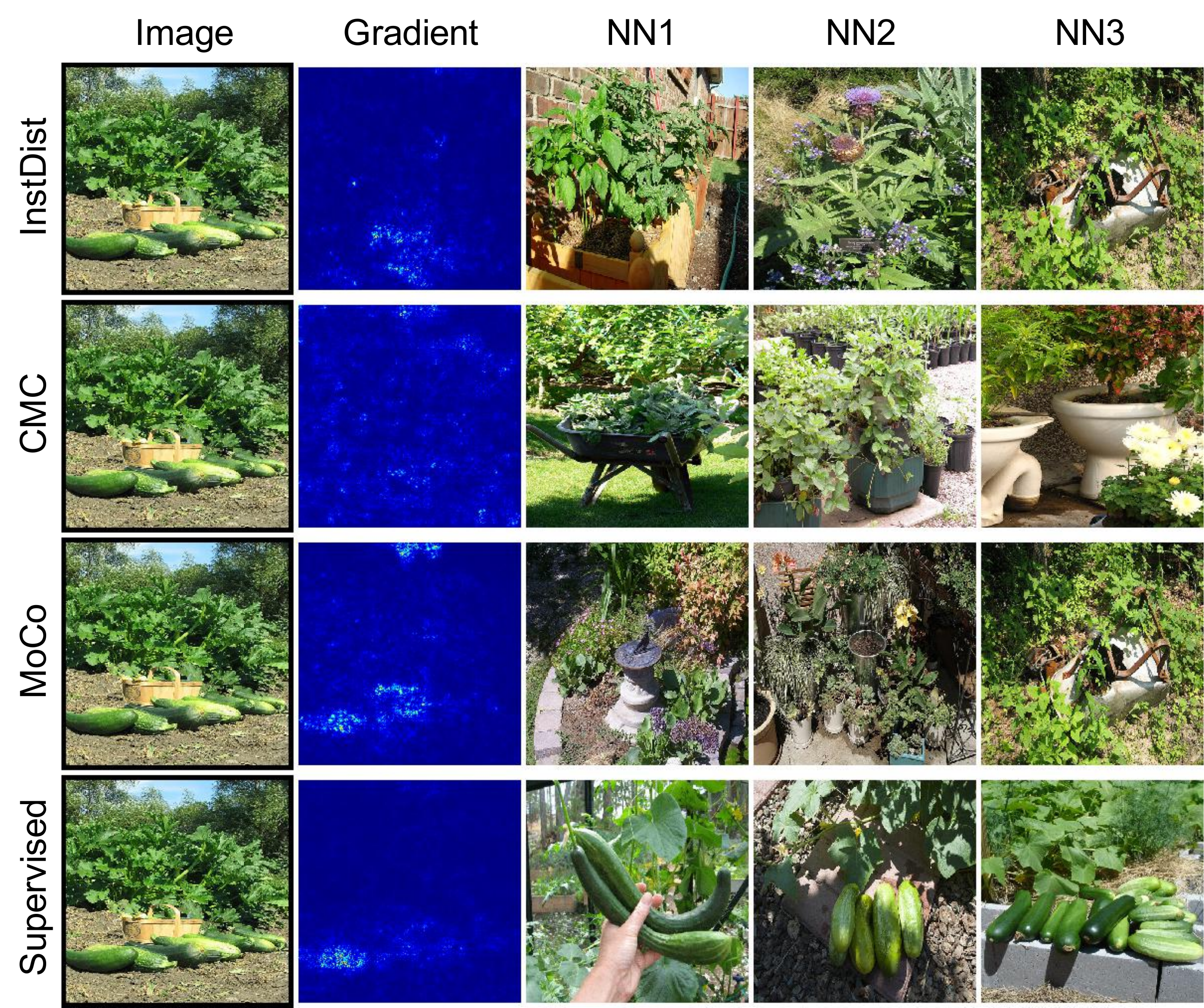}
 	\includegraphics[width=\linewidth]{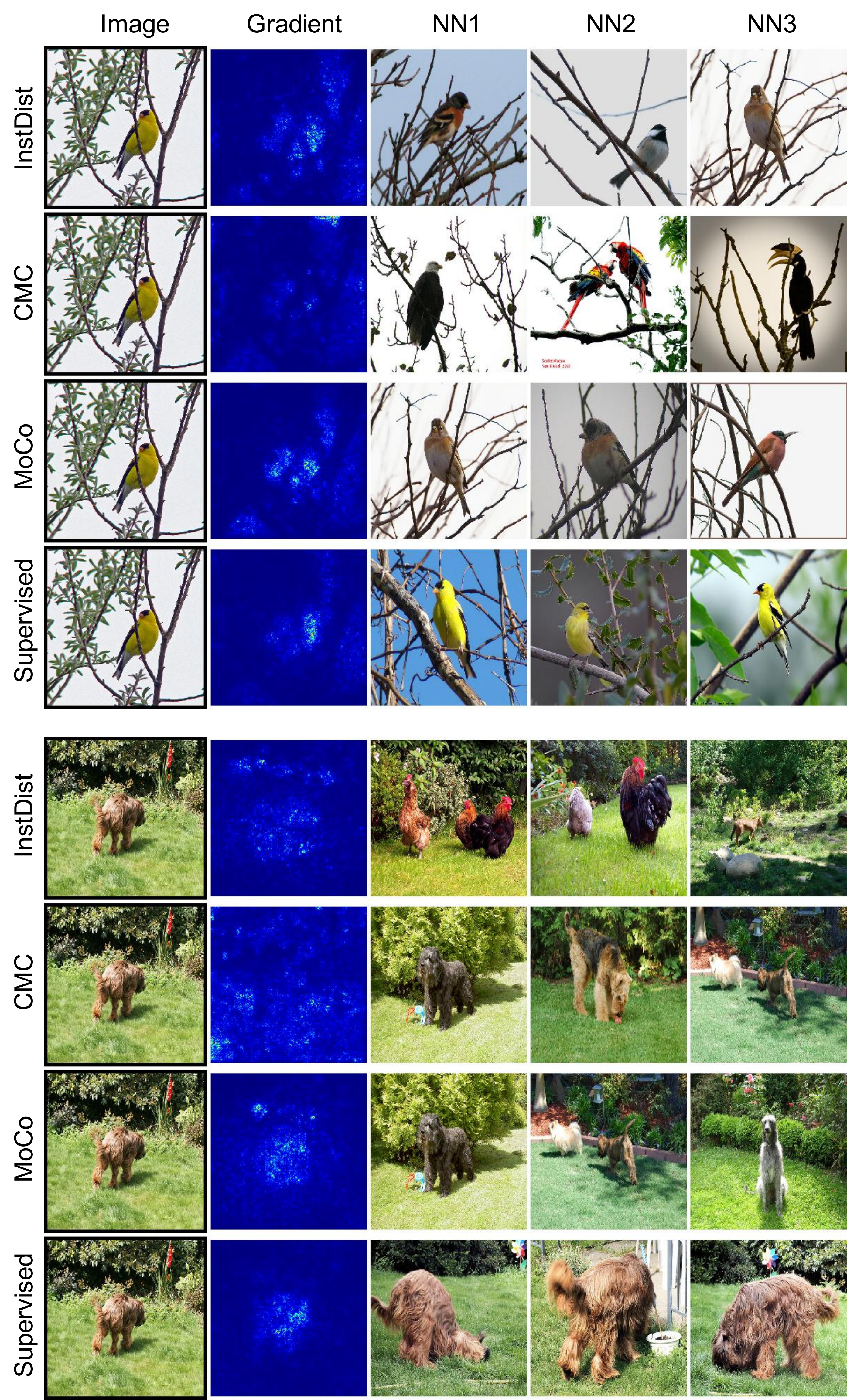}
 	\label{fig:vis0}
 \end{figure}

\section{Visualizing and Diagnosing Self-Supervision}

We provide more visualization examples in the following figures. Given an input for each model, we visualize its top-3 nearest neighbors in the embedding space, as well as the gradient on the pixel space with respect to the classification signal. Compared with the supervised model, which is able to localize the target objects, self-supervised contrastive models (InstDisc, CMC, MoCo) look holistically over the image and are prone to distraction by backgrounds. 
\vspace{3pt}

\begin{figure*}[h]
	\centering
0	\includegraphics[width=0.95\linewidth]{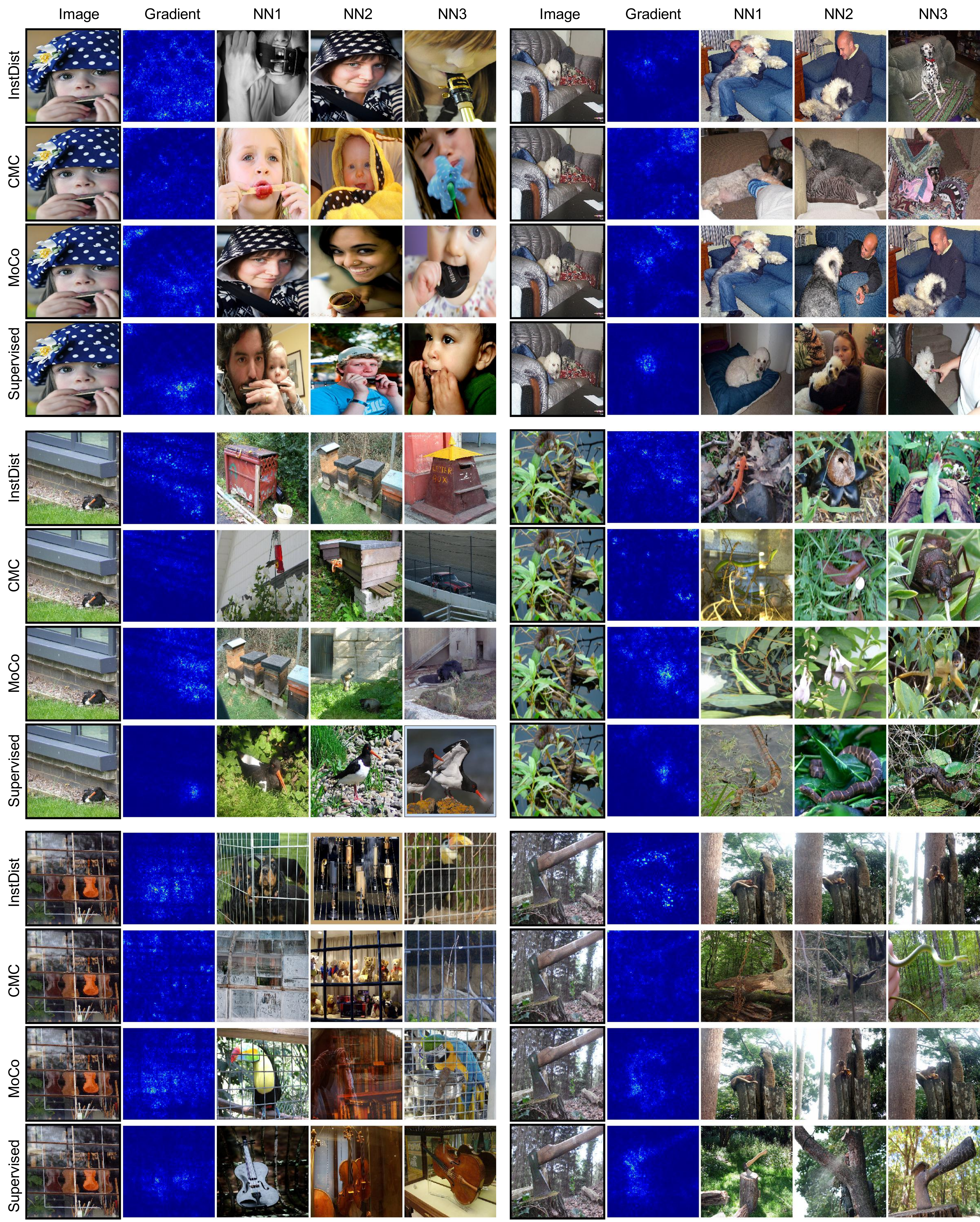}
	\label{fig:vis2}
\end{figure*}

\begin{figure*}[h]
	\centering
	\includegraphics[width=0.95\linewidth]{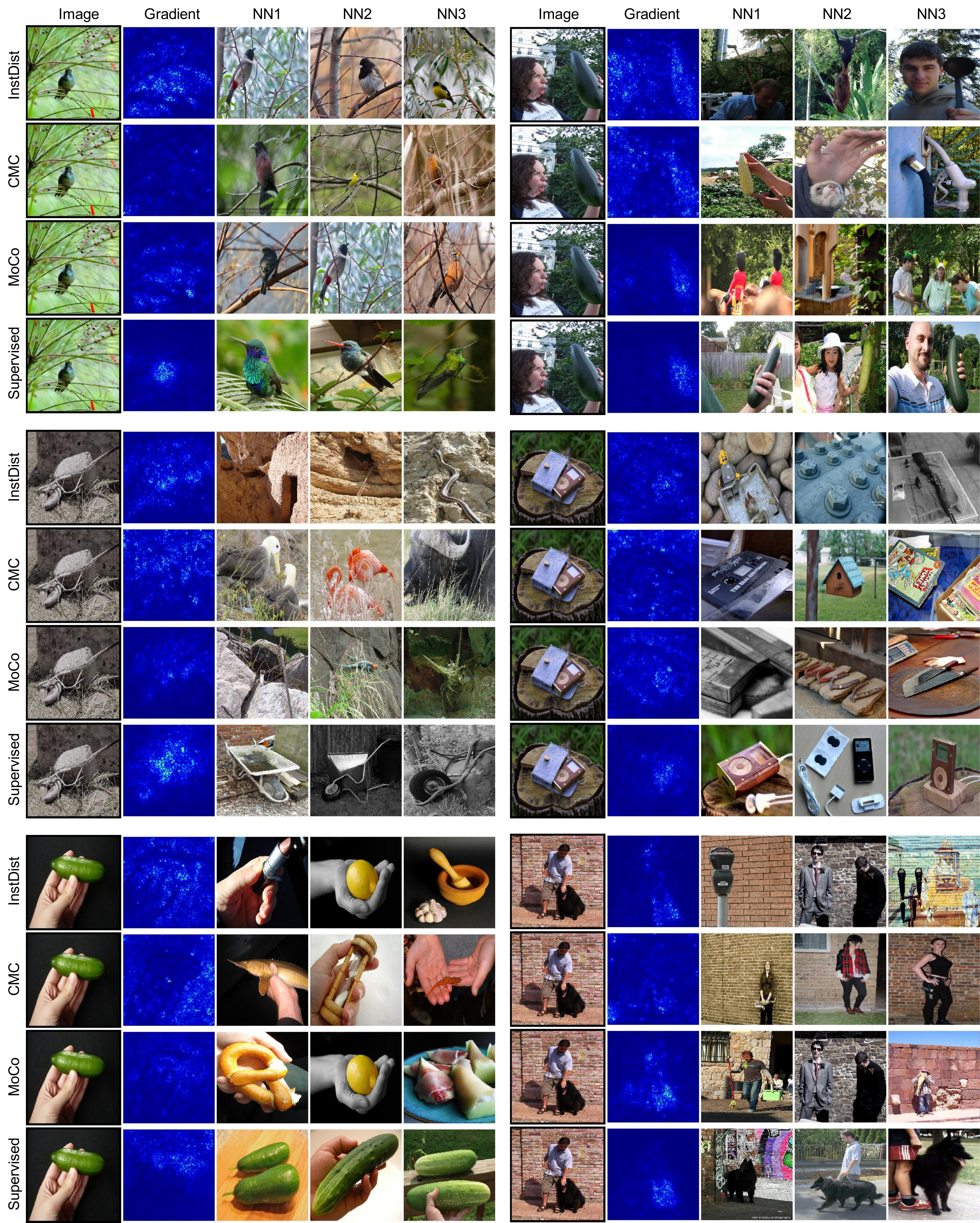}
	\label{fig:vis3}
\end{figure*}

\end{document}